\documentclass[runningheads]{llncs}
\usepackage[T1]{fontenc}
\usepackage{graphicx}
\begin{document}
\title{YCB-LUMA: YCB Object Dataset with Luminance Keying for Object Localization}
\titlerunning{YCB-LUMA for Object Localization}
%

\author{Thomas Pöllabauer\inst{1,2}\orcidID{0000-0003-0075-1181}
}
\authorrunning{Thomas Pöllabauer}
\institute{Fraunhofer Institute for Computer Graphics Research, Darmstadt, Germany \and
Technical University Darmstadt, Darmstadt, Germany
\email{thomas.poellabauer@igd.fraunhofer.de}}
\maketitle              
\begin{abstract}
Localizing target objects in images is an important task in computer vision. Often it is the first step towards solving a variety of applications in autonomous driving, maintenance, quality insurance, robotics, and augmented reality. Best in class solutions for this task rely on deep neural networks, which require a set of representative training data for best performance. Creating sets of sufficient quality, variety, and size is often difficult, error prone, and expensive. This is where the method of luminance keying \cite{pollabauer2024fast,LUMA++} can help: it provides a simple yet effective solution to record high quality data for training object detection and segmentation. We extend previous work that presented luminance keying on the common YCB-V set of household objects \cite{posecnn} by recording the remaining objects of the YCB superset. The additional variety of objects - addition of transparency, multiple color variations, non-rigid objects - further demonstrates the usefulness of luminance keying and might be used to test the applicability of the approach on new 2D object detection and segmentation algorithms.

\keywords{Computer Vision  \and Machine Learning \and Object Localization \and Object Detection \and Segmentation \and Synthetic Data Generation.}
\end{abstract}
\section{Introduction}
Deep learning-based algorithms dominate the task of object detection delivering state-of-the-art accuracy, but relying on the availability of large volumes of high-quality annotated training data. As highlighted in previous work, Deep Neural Networks (DNNs) demand substantial annotated datasets to achieve optimal performance. Traditionally, this data is obtained through manual labeling, which is both error-prone and time-consuming, or through rendering, which necessitates detailed geometry and material information. These methods pose significant challenges, particularly for small-scale applications where resources and time are limited, making it uneconomical to generate the required data.

A streamlined and efficient method for acquiring high-quality training data can significantly broaden the applicability of deep learning techniques, even for niche applications. While chroma keying has been a popular technique for background replacement in image processing, it often suffers from issues such as color bleeding and overlap. Instead, luminance keying, utilizing a black screen with high light absorption to capture videos of target objects, circumvents the typical challenges of chroma keying, allowing for automatic object masking through brightness thresholding, thereby eliminating the need for labor-intensive manual annotation. The technique not only competes favorably with traditional methods like rendering but does so without the need for intricate 3D models or material data, and in a fraction of the time. 

\begin{figure}[t!]
    \centering
    \includegraphics[width=\textwidth]{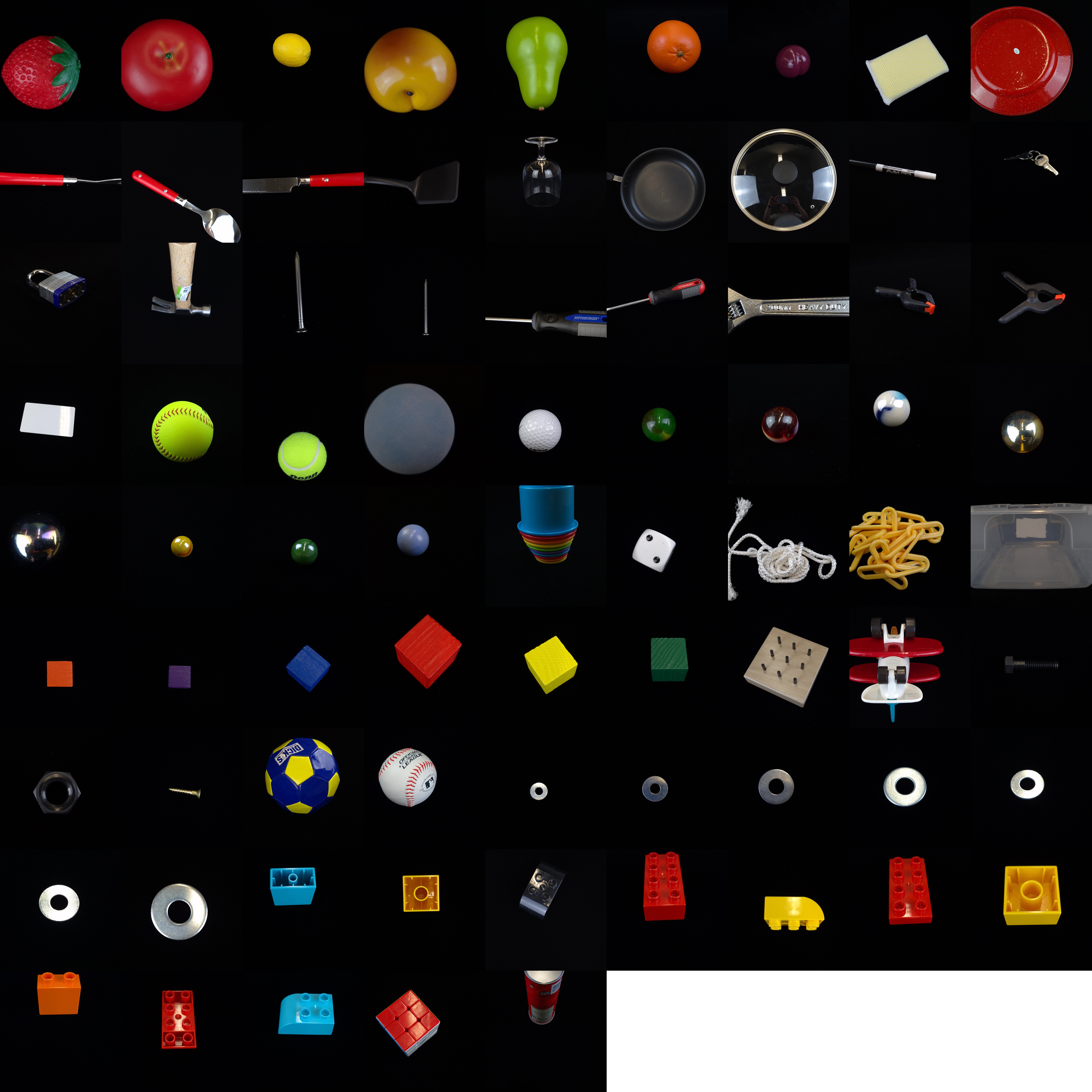}
    \caption{Additional objects to complement the YCB-V set. For each object, we took multiple recordings to capture it from all sides. For deformable objects, such as the yellow chain, we record multiple different states of deformation.}
    \label{fig:overview}
\end{figure}

We extend upon the previous work by complementing the provided subset of objects, YCB-V, with the remaining objects that make up the complete YCB set. We provide the high quality recordings and code for automatic generation of training data and annotations. The extended set significantly improves the variety of the dataset, increasing the meaningfulness of performance validation of tested algorithms.

\section{Related Work}
The YCB object set has proven to be a valuable contribution to the field. Especially its subset YCB-V is of great importance to the evaluation of the 6 degrees of freedom pose estimation problem and has been included as a "classic core dataset" into the Benchmark of Pose Estimation algorithms (BOP) \cite{bop18,bop20,bop23,bop22}.

Additional efforts were invested to extend the dataset to evaluate additional modalities such as neuromorphic cameras \cite{rojtberg2024ycb,rojtberg2023ycb}, stereo vision \cite{pöllabauer2024extending}, and multi-camera setups \cite{ycbm}.

Luminance keying as described in \cite{pollabauer2024fast} follows the tradition of previous methods, placing the target objects in front of different colors used for keying, such as white, green, or gray \cite{grundhofer2010color,lecun2004learning,yamashita2008every}, or more complex backgrounds such as checkerboards \cite{agata2007chroma}, extracting the objects from the background and - using image augmentation methods - use the data to fit the target algorithm. Compared to previous methods focusing on color, luminance keying proved to be simpler to use while using the data to train 2D detectors outperformed chroma keying using a typical green screen \cite{pollabauer2024fast}. 

\section{Methodology}
\begin{figure}
    \centering
    \includegraphics[height=\textheight]{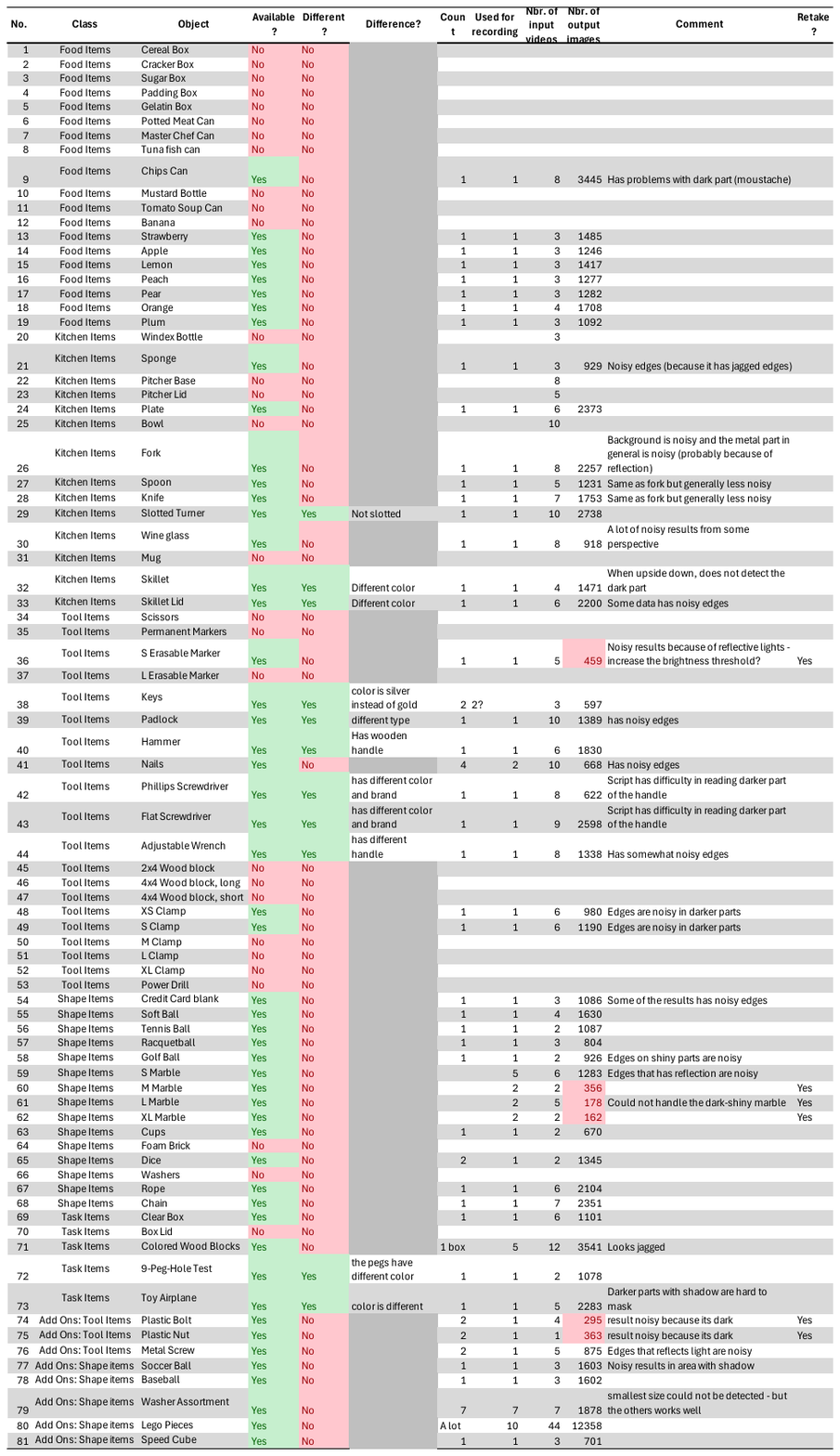}
    \caption{Description of our recordings. Each object is attributed to a category and given a name according to the original publication where possible. We also report whether the object is part of this recording session or not (the YCB-V objects have been recorded previously), whether the recorded objects differ in some way compared to the original YCB data, and some additional meaningful meta data.}
    \label{fig:meta}
\end{figure}

Following the method of data recording with luminance keying, utilizing a 99.99\% light absorbing background \cite{pollabauer2024fast}, we record the additional objects as found in the YCB dataset. These new recordings complement the original ones depicting all of the objects to be found in the YCB-V subset. All of the newly recorded objects are represented in Figure \ref{fig:overview}. Additional relevant meta data is presented in Figure \ref{fig:meta}.

In contrast to the YCB-V subset, the new objects contain a wider range of appearances: there are now multiple transparent objects, such as objects made of transparent plastics and others made of glass, more metallic surfaces, such as on the fork, knife, and spoon, multiple alterations of the same objects, such as the same lego pieces, but of different colors, as well as deformable objects, such as the yellow chain and the white piece of cord. Whenever there are multiple variations of a target object, we differentiate between them by creating separate subfolders. This allows to test for generalization, for instance, by training on one variation and testing on another, or to combine all variations within the training or test set, if required.

For easy processing of these and other objects, which have been recorded following the same setup, we provide some scripts to automatically extract training data for 2D object detector training from the recordings.

\section{Discussion}
We extended the previously available recordings of the YCB-V object dataset by the remaining YCB objects, following the luminance keying approach. Our data allows for performance evaluation of 2D object detectors and segmentation algorithms. We provide our recordings, together with processing code for automated masking, at \url{https://huggingface.co/datasets/tpoellabauer/YCB-LUMA} and \url{https://github.com/tpoellabauer/ycb-luma}. \newline If you are interested in the full YCB dataset, please also download the YCB-V subset from \url{https://huggingface.co/datasets/tpoellabauer/YCB-V-LUMA}.

%
%
%
\bibliographystyle{splncs04}
\bibliography{bib}

\end{document}